\documentclass{styles/svproc}

\usepackage{url}
\usepackage{amsmath} % assumes amsmath package installed
\usepackage{amssymb}  % assumes amsmath package installed
\usepackage{bbm}
\usepackage{cite}
\usepackage[caption=false]{subfig}     % caption=false to avoid conflict with IEEE caption format
\usepackage{enumitem}  
\usepackage{float}
\usepackage{afterpage}
\usepackage{stfloats}
\usepackage{tabularx}
\usepackage{array}
\usepackage{makecell}
\usepackage[nosetpagesize]{graphicx}  % nosetpagesize is critical to maintaining the page size 
\usepackage{booktabs}
\usepackage{multirow}
\usepackage{dsfont}
\usepackage{romannum}
\usepackage{xstring}
\usepackage{array}
\usepackage{subcaption}
\usepackage{titlecaps}

\usepackage{graphicx,wrapfig,lipsum}
\usepackage{bold-extra}

\Addlcwords{w/o}     % This is great -- adding exceptions to capitalization rules 
 
\renewcommand{\eqref}[1]{(\ref{#1})}
\newcommand{\figref}[1]{Fig.~\ref{#1}}
\newcommand{\subfig}[1]{\textit{#1}}
\newcommand{\tabref}[1]{Table~\ref{#1}}

\newcommand{\eqnref}[1]{Eq.~\eqref{#1}}

\newcommand{\envref}[1]{\textsc{Env#1}}

\newcommand{\theoref}[1]{Theorem~\ref{#1}}

\newcommand{\ie}{\textrm{i.e.}}
\newcommand{\eg}{\textrm{e.g.}}

\def\rgb{RGB}
\def\gan{BDA-Real2sim}
\def\maxcover{MaxCover}
\def\ours{SEER}  % {\scshape Seer}
\def\nonav{\ours{} w/o Sem2nav}
\def\nodepth{\ours{} w/o Depth}
\def\nosemantics{\ours{} w/o Semantics}

\def\semseg{semantics}
\def\keypoints{2D keypoints}
\def\normal{surface normal}
\def\depth{depth}
\def\navigability{navigability}
\def\unsupseg{2D Segmentation}

\def\SR{SR}
\def\SPL{SPL}

\def\turnleft{\texttt{turn\_left}}
\def\turnright{\texttt{turn\_right}}
\def\goforward{\texttt{go\_forward}}

\newcommand{\hypo}[1]{H\textsubscript{#1}}
\def\h{\hypo{1}}
\def\hh{\hypo{2}}
\def\hhh{\hypo{3}}
\def\hhhh{\hypo{4}}

\def\adist{$\mathcal{A}$-distance}
\def\X{\mathcal{X}}
\def\Y{\mathcal{Y}}
\def\Z{\mathcal{Z}}
\def\D{\mathcal{D}}
\def\S{\mathcal{S}}
\def\H{\mathcal{H}}
\DeclareMathOperator*{\E}{\mathbb{E}}

\DeclareMathOperator*{\argmin}{arg\,min}

\def\weblink{\urllink[pre = \bgroup\bf, post = \egroup]}

\begin{document}
\mainmatter              % start of a contribution

\title{Invariance is Key to Generalization: Examining the Role of Representation in Sim-to-Real Transfer for Visual Navigation\vspace{-5pt}} % \vspace{-10pt}

% SEER: Spatial-Semantic Representation for Sim-to-Real Transfer in Visual Navigation

% \title{Invariance is Key to Generalization: Reducing Domain Gap with Invariant Representations in Sim-to-Real Transfer\vspace{-10pt}} 

\titlerunning{Generalization through Invariance}  % abbreviated title (for running head)
\author{Bo Ai \and Zhanxin Wu \and David Hsu} %  \vspace{-5pt}

\authorrunning{B. Ai et al.} % abbreviated author list (for running head)
% \tocauthor{Ivar Ekeland, Roger Temam, Jeffrey Dean, David Grove,
% Craig Chambers, Kim B. Bruce, and Elisa Bertino}
%
\institute{National University of Singapore\\
Singapore 119077, Singapore\\
\email{bo.ai@u.nus.edu}, \email{zhanxinwu@u.nus.edu}, \email{dyhsu@comp.nus.edu.sg}}

% \\ WWW home page:
% \texttt{http://users/\homedir iekeland/web/welcome.html}
% \and
% Universit\'{e} de Paris-Sud,
% Laboratoire d'Analyse Num\'{e}rique, B\^{a}timent 425,\\
% F-91405 Orsay Cedex, France}

\maketitle              % typeset the title of the contribution
\vspace{-10pt}

\begin{abstract}
The data-driven approach to robot control has been gathering pace rapidly, yet generalization to unseen task domains remains a critical challenge. We argue that the key to generalization is representations that are (i) rich enough to capture all task-relevant information and (ii) invariant to superfluous variability between the training and the test domains. We experimentally study such a representation---containing both depth and semantic information---for visual navigation and show that it enables a control policy trained entirely in simulated indoor scenes to generalize to diverse real-world environments, both indoors and outdoors. Further, we show that our representation reduces the \adist{} between the training and test domains, improving the generalization error bound as a result. Our proposed approach is scalable: the learned policy improves continuously, as the foundation models that it exploits absorb more diverse data during pre-training.

% The abstract should summarize the contents of the paper
% using at least 70 and at most 150 words. It will be set in 9-point
% font size and be inset 1.0 cm from the right and left margins.
% There will be two blank lines before and after the Abstract. \dots
% We would like to encourage you to list your keywords within
% the abstract section using the \keywords{...} command.

\end{abstract}

\section{Introduction}

\begin{wrapfigure}{r}{0.5\textwidth}
    \vspace{-20pt}
    \includegraphics[width=0.5\textwidth]{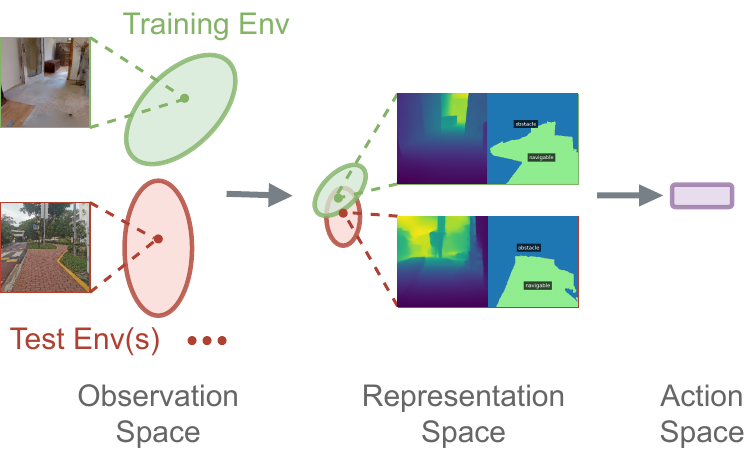}
    \caption{Transforming different domains into a compact representation space reduces the generalization gap between training and test environments. }
    \label{fig:mapping}
    \vspace{-20pt}
\end{wrapfigure} 

Recent years have witnessed the increasing popularity of robot learning, yet data scarcity and generalization remain critical challenges. While simulators may serve as ``data factories'', learned policies often face performance degradation, because of the \textit{sim-to-real} gap, \ie, the data distribution shift between the simulated and the real world \cite{challenges_embodied}. For reliable robot performance, robustness against such domain shifts is essential. Our work studies one mechanism that enables learned policies to generalize across domains (\figref{fig:mapping}).

We target the task of visual navigation. Given a goal location, the robot receives an RGB image as observation at each time step and then predicts a motion command for actuation. Our objective is to learn a policy entirely in simulation, and enable it to generalize to the real world. This is challenging, because the high-dimensional input space, consisting of all possible visual images, induces enormous variability. Our idea is to inject inductive bias into the learning system~\cite{foundation_learning}, in particular, a structured representation. 

How do we acquire such a representation? It should fulfill two conditions. First, it captures sufficient information for the end task objective, \eg, predicting control commands. Second, to be robust against domain shifts, it contains little extraneous information and has low information entropy. We expect such a representation to be \textit{invariant} across domains. 
% so that it is stable and consistent across domains.  We call such representations as \textbf{invariant representations}. 
Clearly, to achieve invariance,  there is a trade-off between representation \textit{informativeness}  and \textit{compactness}.

%%%
\begin{figure}[t]
  \centering
  \begin{tabular}{c@{\hspace{12pt}}c}  % height 0.17 0.21
      \includegraphics[height=0.195\columnwidth]{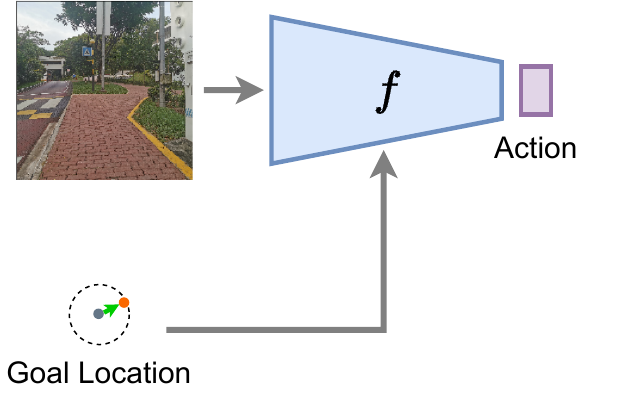} &
      \includegraphics[height=0.245\columnwidth]{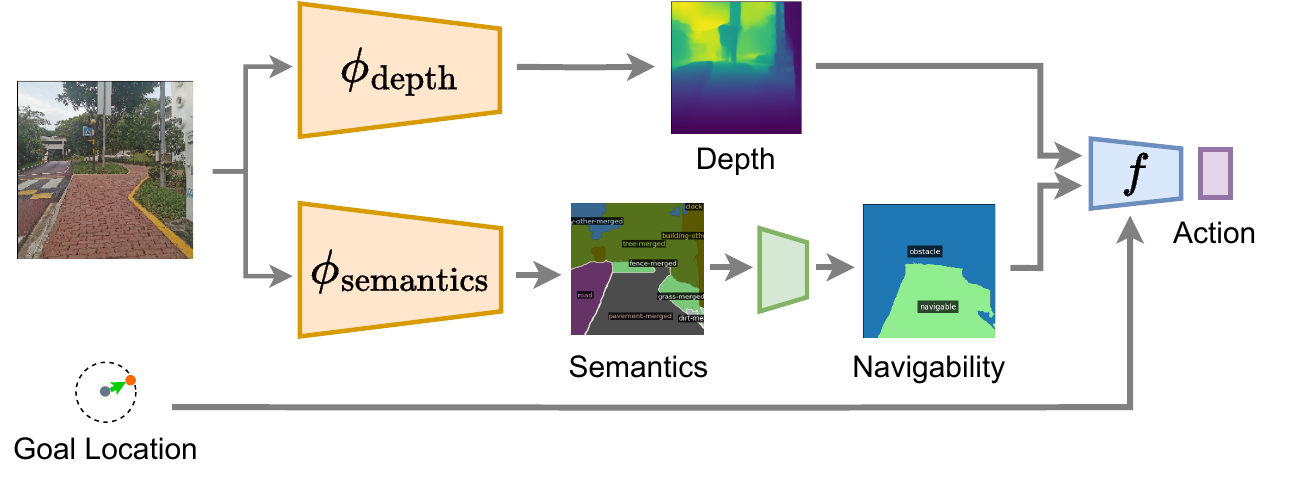} \\
      (\subfig{a}) & (\subfig{b}) \\
  \end{tabular}  % \setlength{\tabcolsep}{1pt} % Default value: 6pt
  \vspace{-10pt}
  \caption{A structured representation as an inductive bias for learning robot control policy $f$.  (\subfig a) The standard end-to-end learning approach (\eg, \cite{decision}) uses an image of raw pixels as the input to $f$. (\subfig b) Our approach uses pre-trained representation models, $\phi_\textrm{depth}$ and $\phi_\textrm{semantics}$,  to extract invariant representations and improve the generalization performance of the learned policy. }
  \label{fig:conceptual_mm2} 
  \vspace{-10pt}
\end{figure}
%%%

We propose an instance of such representation for visual navigation: {\bfseries \ours{}}, standing for \textbf{S}patial s\textbf{E}mantic r\textbf{E}p\textbf{R}esentation, consists of two features: \depth{} and \navigability{} (\figref{fig:conceptual_mm2}). We extract features with pre-trained vision models, learn the mapping from features to control commands in simulation, and test the controller in diverse real-world environments. Our experimental results show that, albeit the simulation is purely indoor environments, our approach enables the policy to generalize effectively to indoor and even outdoor scenes. This significantly outperforms some of the best methods for sim-to-real transfer. We also provide additional experimental evidence that indicates a reduced domain gap in the representation space and thus explains the improved generalizability.

\section{Related Work}

Generalization is a fundamental question in machine learning and has a vast literature. Within robot learning, for sim-to-real transfer, the two most common approaches are domain randomization (DR) and domain adaptation (DA).

The main idea behind DR is to randomize the training domain so that the test domain appears as one variation of training. DR has been effective in robotic manipulation \cite{openai-dexterous}, but has shown limited success in navigation because of the difficulty of achieving sufficient data coverage over the diverse open-world environments in which a mobile robot might operate. On the other hand, DA seeks to learn a mapping from one domain to the other so that the observations in the training and test domains appear similar. However, learning such a mapping requires a large amount of data from both domains and the optimization is often difficult. Both VR-Goggles \cite{vr-goggles} and BDA \cite{da-bidirec} use CycleGAN \cite{cycle-gan} to transform visual observations at test time back to the training domain, but CycleGAN is known to be weak at modeling geometric transformations \cite{cycle-gan}, limiting the applicability of the sim-to-real approach. 
% To be specific, there has not been prior work that demonstrated transferring simulated indoor-learned policy to real outdoor environments, though the underlying control mechanism of navigation should be similar in different domains, \ie, identifying navigable regions and avoiding obstacles. 

In this work, we believe that injecting invariance is a scalable and practical way of achieving generalization \cite{foundation_learning}. Our approach is related to the work on mid-level representations \cite{midlevel-navigation, midlevel-manipulation}. The earlier work proposes a maximum coverage feature set comprised of four features (\ie, surface normal, 2D keypoints, 2D segmentation, semantic segmentation) and shows it has significant information overlap with all 26 features in \cite{taskonomy} and leads to good generalization performance for control policies. 
% Our work narrows down the feature set to just two features and shows that the reduced feature set provides better generalization and efficiency for visual navigation.
Our work proposes a novel set of only two features, demonstrating that this compact feature set provides much better generalization for visual navigation.
% We provide the first study that benchmarks both invariance-based approaches and DA approaches in diverse challenging real-world scenes. (iii) 
Further, we connect the empirical performance to theories on generalization~\cite{generalization_theory} and point out that the gap between different environments is reduced in our representation space, validating our approach and gaining a theoretical understanding for further investigation.

\section{Technical Approach}

% \subsection{Principle: The Informativeness-Compactness Trade-Off} \label{sect:principle}
% \vspace{-5pt}

% \begin{figure}[t]
%   \centering{\includegraphics[width=0.55\columnwidth]{method/representation-bottleneck.pdf}}
%   \vspace{-3pt}
%   \caption{Trade-off between compactness and informativeness. $o_t$ is the RGB observation, $\phi$ is a feature extractor module, $f$ is the mapping from intermediate representation $\phi(o_t)$ to the final action. The composition of $\phi$ and $f$ is the policy, \ie, $\pi = f \circ \phi$. The trade-off is that we want a smaller end $a$ for compactness but a larger end $b$ for more information for the end task prediction.}
%   \label{fig:compactness}
%   \vspace{-10pt}
% \end{figure} 

To enable a policy to generalize across domains, the representation could act as an abstraction that filters out task-irrelevant details. The more compact a representation is, the more it is invariant to visual observation $o_t$ from different distributions, whereas the less information it could potentially provide to the downstream module $f$. 

To formalize this intuition, we use discrete random variables $\X$, $\Y$, and $\Z{}$ to represent observation, representation, and the predicted controls respectively. Since each variable predicts the subsequent one, $\X \rightarrow \Y \rightarrow \Z$ forms a Markov chain, and the model's predictive uncertainty can be decomposed and bounded by the cardinality of the objective and some information entropy terms: 
\begin{align*}
        H(\Z|\X) 
        & = \E \left[ \log p(\Z|\X) \right] \\
        & = \E \left[ \log p(\Z|\Y) + \log p(\Y|\X) \right] \\
        & = H(\Z|\Y) + H(\Y|\X) \\
        & = H(\Z) - I(\Z;\Y) + H(\Y|\X) \\
        & \leq H(\Z) - I(\Z;Y) + H(\Y) \\
        & \leq \log({|\Z|}) - I(\Z;\Y) + H(\Y)
\end{align*}
where $I(\Z;\Y)$ denotes the mutual information between $\Z$ and $\Y$, and ${|\Z|}$ is the cardinality of $\Z$. The inequality shows that, to make a model make confident predictions, \ie, to minimize $H(\Z|\X)$, we can minimize $H(\Y)$, the information contained in the representation, and maximize $I(\Z;\Y)$, the relevance of information in representation with respect to the final objective. Thus, these two terms conceptualize compactness and informativeness respectively. It is worth noting that $H(\Y)$ is bounded by $\log({|\Y|})$, thus it is a useful heuristic in practice that we use representations that lie in low-dimensional space. 

% \subsection{Proposed Representation} \vspace{-5pt}
In this work, we introduce one instance of such representation for navigation. We conjecture that the world can be decomposed into geometry and semantics, both are useful to predict the control commands: 
\begin{itemize} \vspace{-5pt}
    \item Geometry: 3D occupancy of points in the space, which encapsulates information about the presence of obstacles, \eg, walls and trees. 
    \item Semantics: Attributes associated with the points in the space, which may capture traversability, \eg, a path is drivable while a wall is an obstacle.
\end{itemize} \vspace{-5pt}
These two pieces of information can be readily extracted with pre-trained depth estimation and semantic segmentation models. Specifically, we find Mask2Former \cite{mask2former} and DPT \cite{dpt} to work well for in-the-wild data without fine-tuning. 

\begin{figure}[t]
  \centering
  \begin{tabular}{cccccc}
     % &  \thead{Simulation } & \thead{Indoor \\ Lab } & \thead{Indoor \\ Corridor } & \thead{Outdoor \\ Pavement } & \thead{Outdoor \\ Grassland } \\ \vspace{2pt}
     &  \thead{\envref{0} \\ Simulation} & \thead{\envref{1} \\ Lab  } & \thead{\envref{2} \\ Corridor} & \thead{\envref{3} \\ Pavement } & \thead{\envref{4} \\ Grassland } \\ \vspace{2pt}
    RGB & 
    \raisebox{-.5\height}{\includegraphics[width=0.16\columnwidth]{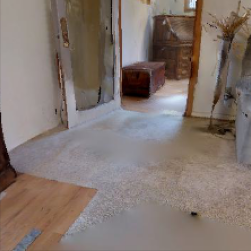}} &
    \raisebox{-.5\height}{\includegraphics[width=0.16\columnwidth]{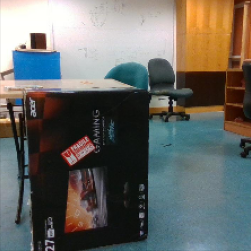}} &
    \raisebox{-.5\height}{\includegraphics[width=0.16\columnwidth]{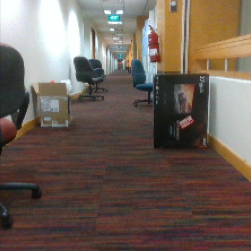}} & 
    \raisebox{-.5\height}{\includegraphics[width=0.16\columnwidth]{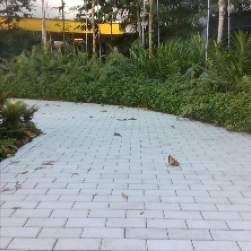}} &
    \raisebox{-.5\height}{\includegraphics[width=0.16\columnwidth]{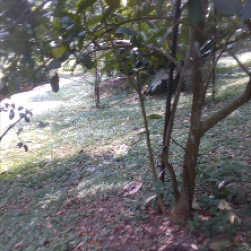}}
    \\ \vspace{2pt}
    Depth & 
    \raisebox{-.5\height}{\includegraphics[width=0.16\columnwidth]{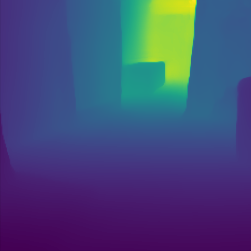}} &
    \raisebox{-.5\height}{\includegraphics[width=0.16\columnwidth]{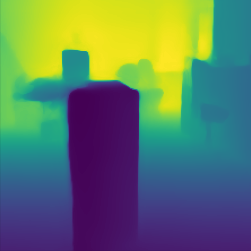}} &
    \raisebox{-.5\height}{\includegraphics[width=0.16\columnwidth]{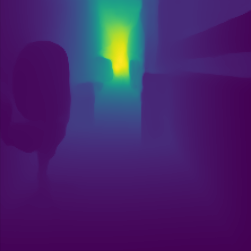}} & 
    \raisebox{-.5\height}{\includegraphics[width=0.16\columnwidth]{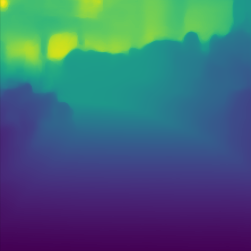}} &
    \raisebox{-.5\height}{\includegraphics[width=0.16\columnwidth]{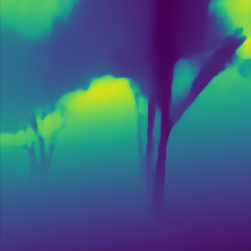}}
    \\ \vspace{2pt}
    Semantics & 
    \raisebox{-.5\height}{\includegraphics[width=0.16\columnwidth]{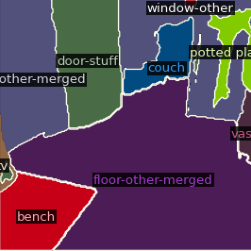}} &
    \raisebox{-.5\height}{\includegraphics[width=0.16\columnwidth]{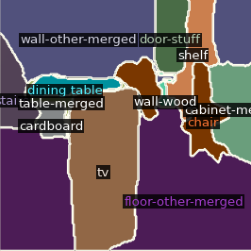}} &
    \raisebox{-.5\height}{\includegraphics[width=0.16\columnwidth]{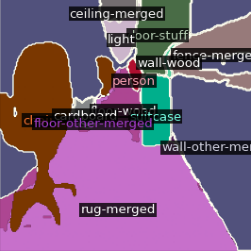}} & 
    \raisebox{-.5\height}{\includegraphics[width=0.16\columnwidth]{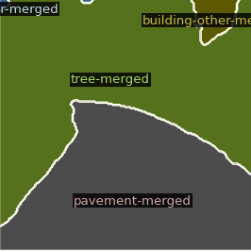}} &
    \raisebox{-.5\height}{\includegraphics[width=0.16\columnwidth]{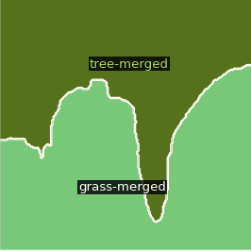}}
    \\ \vspace{2pt}
    Navigability & 
    \raisebox{-.5\height}{\includegraphics[width=0.16\columnwidth]{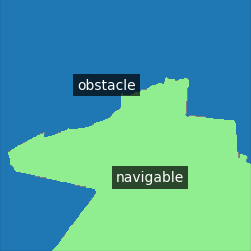}} &
    \raisebox{-.5\height}{\includegraphics[width=0.16\columnwidth]{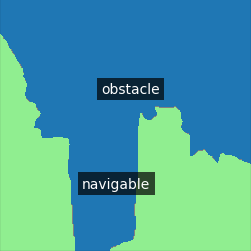}} &
    \raisebox{-.5\height}{\includegraphics[width=0.16\columnwidth]{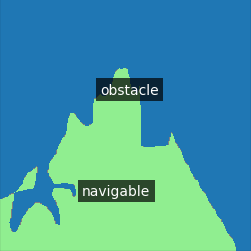}} & 
    \raisebox{-.5\height}{\includegraphics[width=0.16\columnwidth]{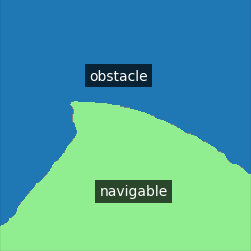}} &
    \raisebox{-.5\height}{\includegraphics[width=0.16\columnwidth]{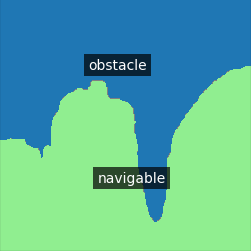}}
    \\
  \end{tabular}
  \vspace{-5pt}
  \caption{Representations during navigation experiments. It can be observed that the environments appear more visually similar in the depth and navigability space. However, with navigability alone, the agent suffers from size-distance ambiguity in geometrically complex environments, such as the grassland, in which case depth information is particularly helpful for obstacle avoidance. }
  \label{fig:exp_obs}
  \vspace{-15pt}
\end{figure}

In addition, we observe that the semantic segmentation mask contains redundant information, \eg, the agent does not need to differentiate between a tree and a wall to avoid a collision. Thus, we make the representation more compact by converting object categories to navigability measures (\tabref{table:seg2nav}). It can be observed that this conversion significantly reduces the domain gap (\figref{fig:exp_obs}).

\begin{table}[t]
\caption{Mapping of object categories to navigability.}
\label{table:seg2nav} 
% \vspace{6pt}
\centering
% \begin{tabular}{p{0.18\columnwidth}  @{\hspace{1cm}}p{0.70\columnwidth}}
\begin{tabular}{p{0.18\columnwidth} @{\hspace{1cm}} p{0.70\columnwidth}}
  \toprule
   \multicolumn{1}{l}{Navigability} & \multicolumn{1}{c}{Object Category} \\
  \hline
  Obstacle & person, bicycle, car, motorcycle, airplane, bus, train, truck, window-other, tree-merged... \\
  \hline
  Ambiguous/ Unrelated  & ceiling-merged, sky-other-merged, mountain-merged, dirt-merged... \\ 
  \hline
  Free space & gravel, playing field, stairs, floor-other-merged, pavement-merged, rug-merged, floor-wood... \\ 
  \bottomrule
\end{tabular}
\vspace{-15pt}
\end{table} 

% \makecell[l]{ceiling-merged, sky-other-merged, \\ mountain-merged, dirt-merged}

% %%%
% \begin{figure}[t]
%   \centering
%   \includegraphics[width=0.9\columnwidth]{method/sim_nav_example1.png} 
%   \includegraphics[width=0.9\columnwidth]{method/real_nav_example1.png} 
%   \caption{Simulation and the real world in the representation space. It can be observed that the two domains are similar in depth space, but they are significantly different in semantic space. Converting semantics to \navigability{} could significantly reduce the gap while only keeping the most important information. }
%   \label{fig:sim-real-gap}
%   % \vspace{-9pt}
% \end{figure}
% %%%

\section{Experiments} \label{experiments}

\subsection{Setup} 
We seek to examine the following hypotheses to show that our method is compact, informative, and effective in enabling sim-to-real transfer.  \vspace{-3pt}
\begin{enumerate}[label=\hypo{\arabic*}., leftmargin=27pt] %[label=\roman*.]
    \item Our representation enables the policy to zero-shot transfer from simulated indoor to real outdoor environments, significantly outperforming GAN-based domain adaptation approaches. 
    \item Reducing semantics categories to \navigability{} is effective in improving the controller's generalizability to semantically different environments.
    \item Our representation is compact. Both geometry and \navigability{} are essential, omitting one of which causes performance degradation.
    \item Our representation is informative to the end task. Injecting more information may not improve task performance. 
\end{enumerate}  \vspace{-5pt}

\begin{figure*}[t]
    \centering
    % \scriptsize
    % \setlength{\tabcolsep}{2pt}  
    \begin{tabular}{c@{\hspace{3pt}}|c@{\hspace{7pt}}c}
    % \scriptsize
        \raisebox{+0.8\height}{\textsf{Training}} & \raisebox{+0.8\height}{\textsf{Testing}}  \\
        % \scriptsize
        \begin{tabular}{c}
            \includegraphics[width=0.3\textwidth]{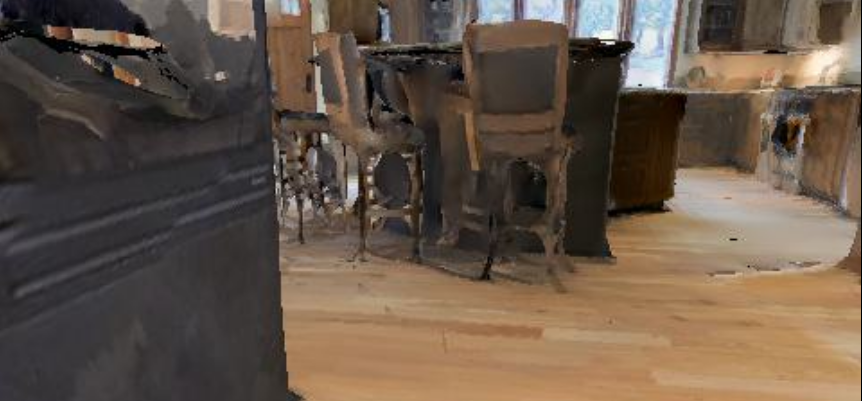} \\
            % (\subfig{0}) Simulation \\ 
            \\
            \includegraphics[width=0.3\textwidth]{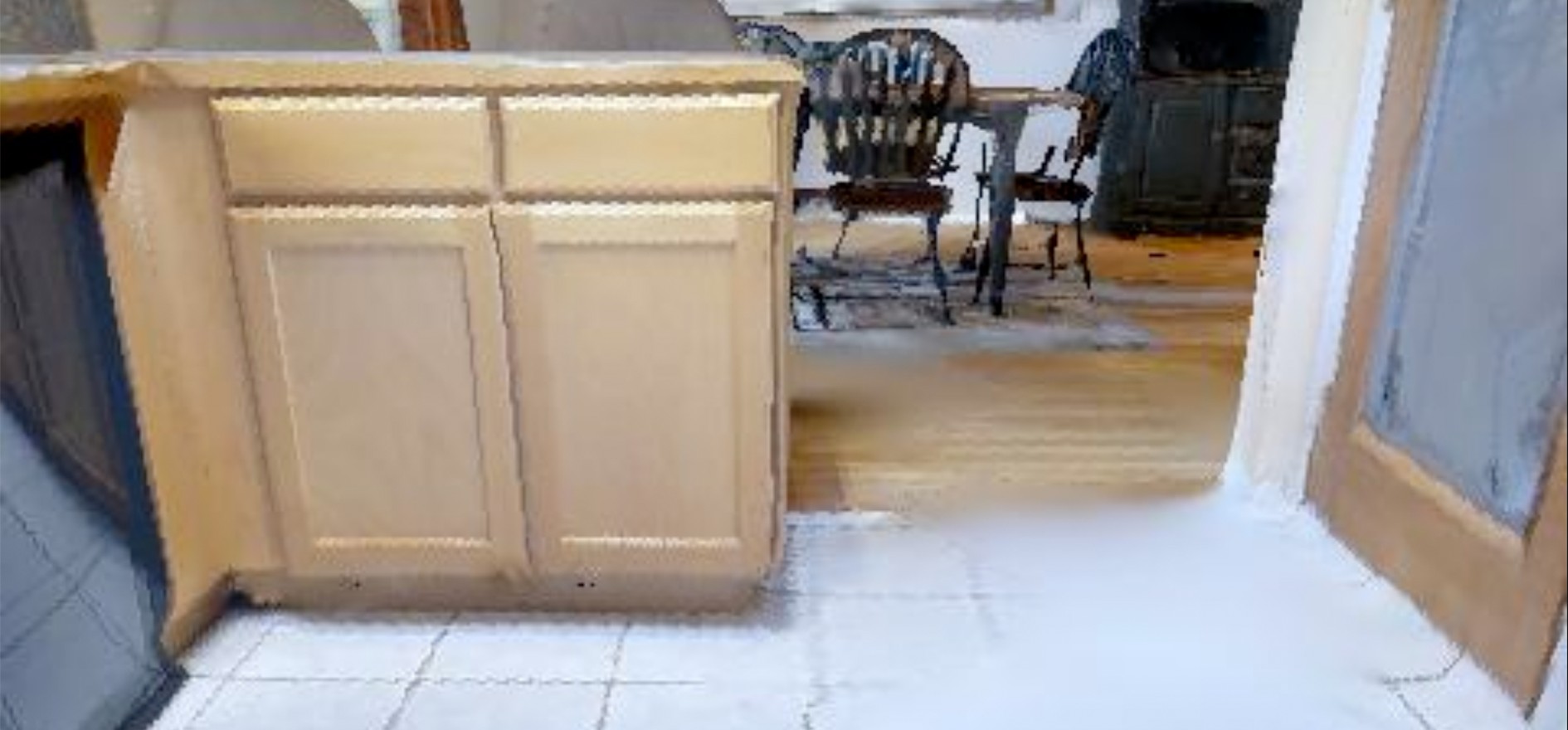} \\
            (\subfig{0}) Indoor Simulation \\ 
        \end{tabular}
      & 
      \hspace{2pt}
      % \scriptsize
      \begin{tabular}{cc}
        \includegraphics[width=0.30\textwidth]{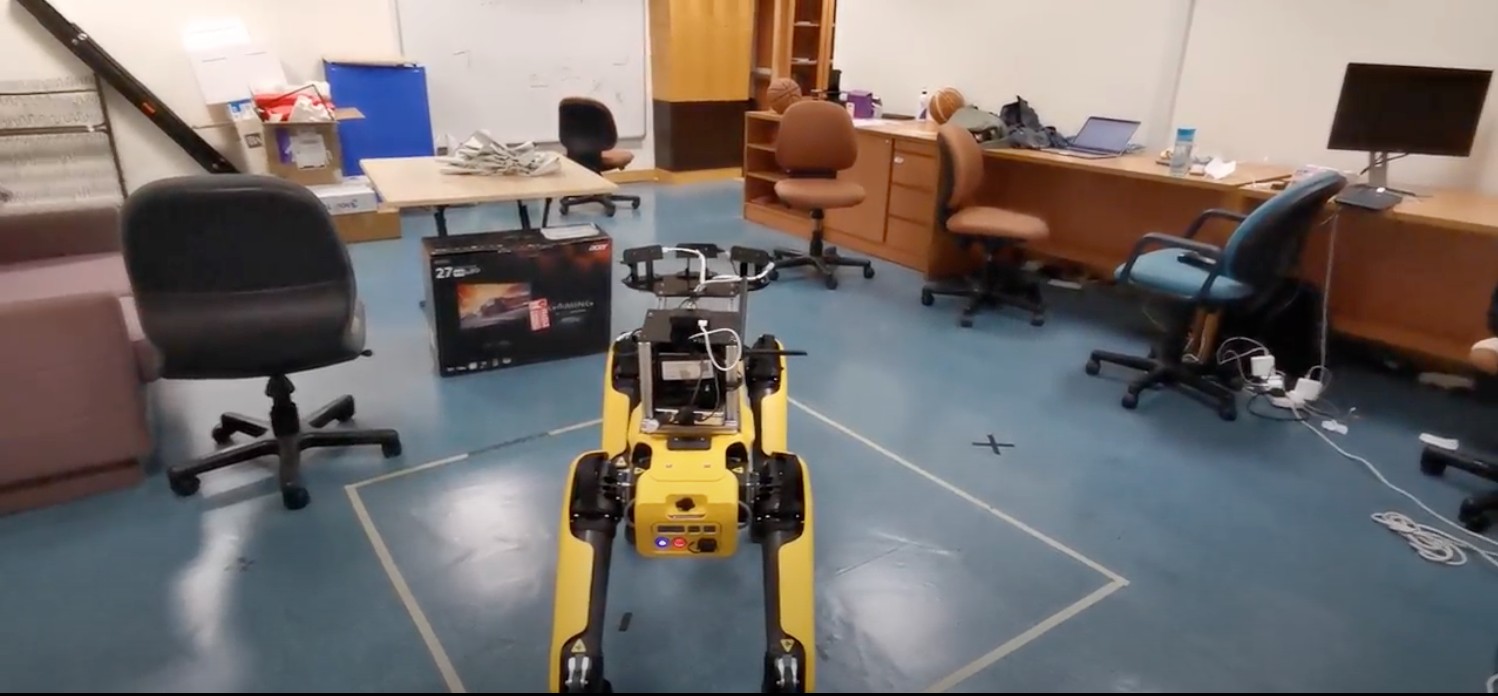}  &
        \includegraphics[width=0.30\textwidth]{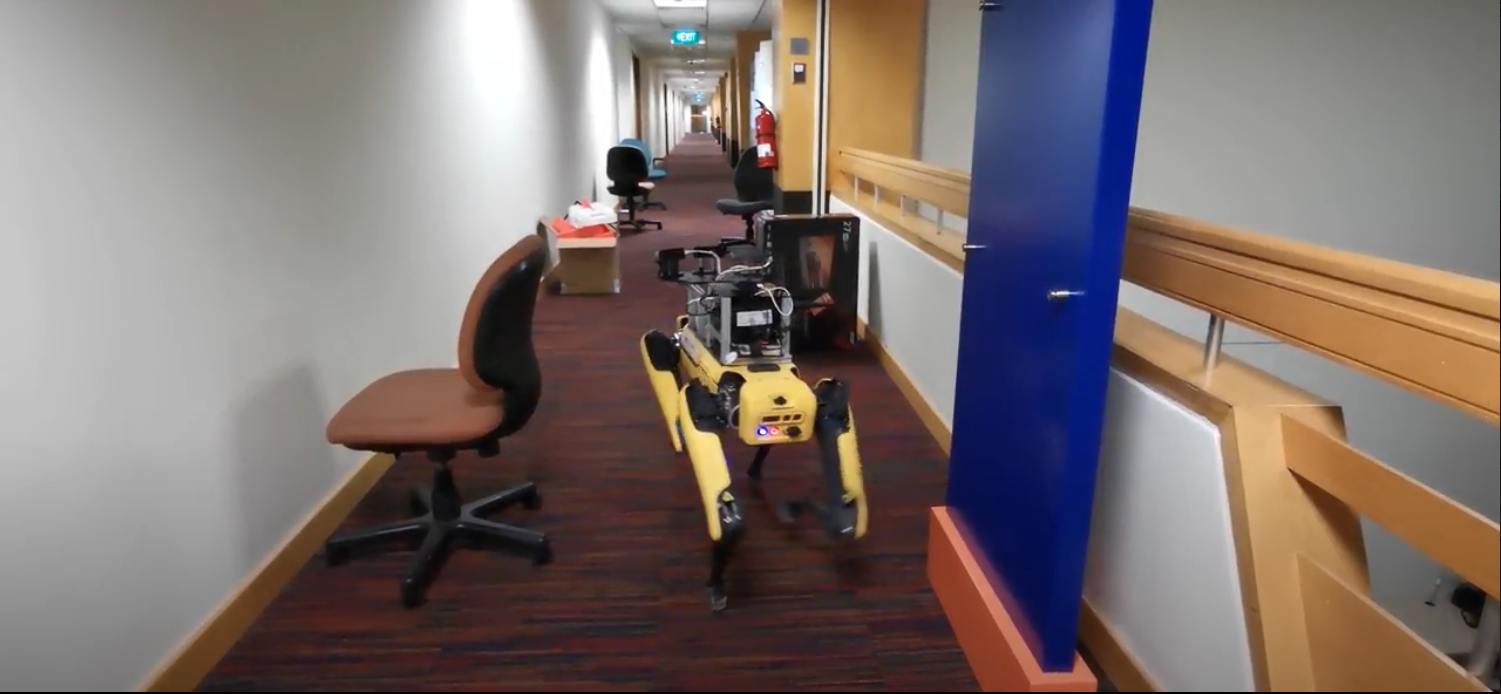}  \\ 
        (\subfig{1}) Indoor Lab & (\subfig{2}) Indoor Corridor \\
         \includegraphics[width=0.30\textwidth]{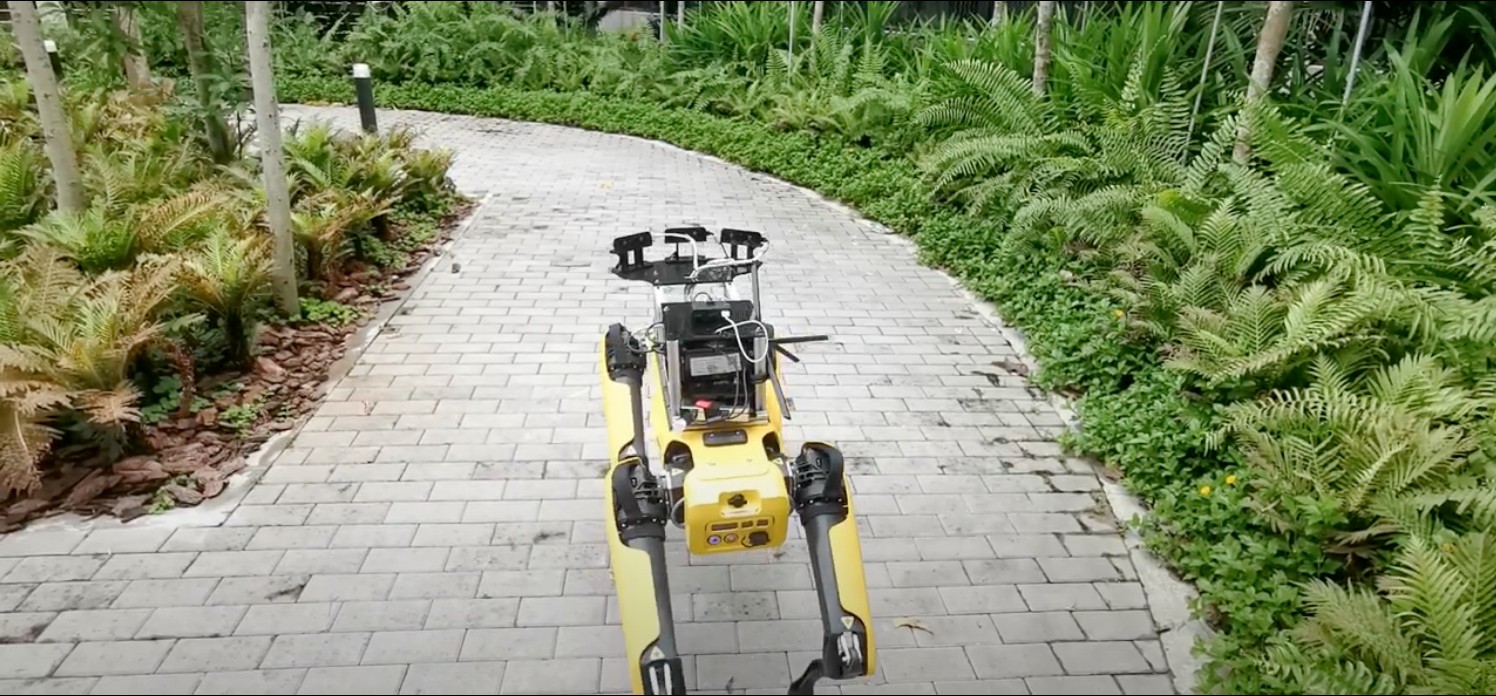} &
         \includegraphics[width=0.30\textwidth]{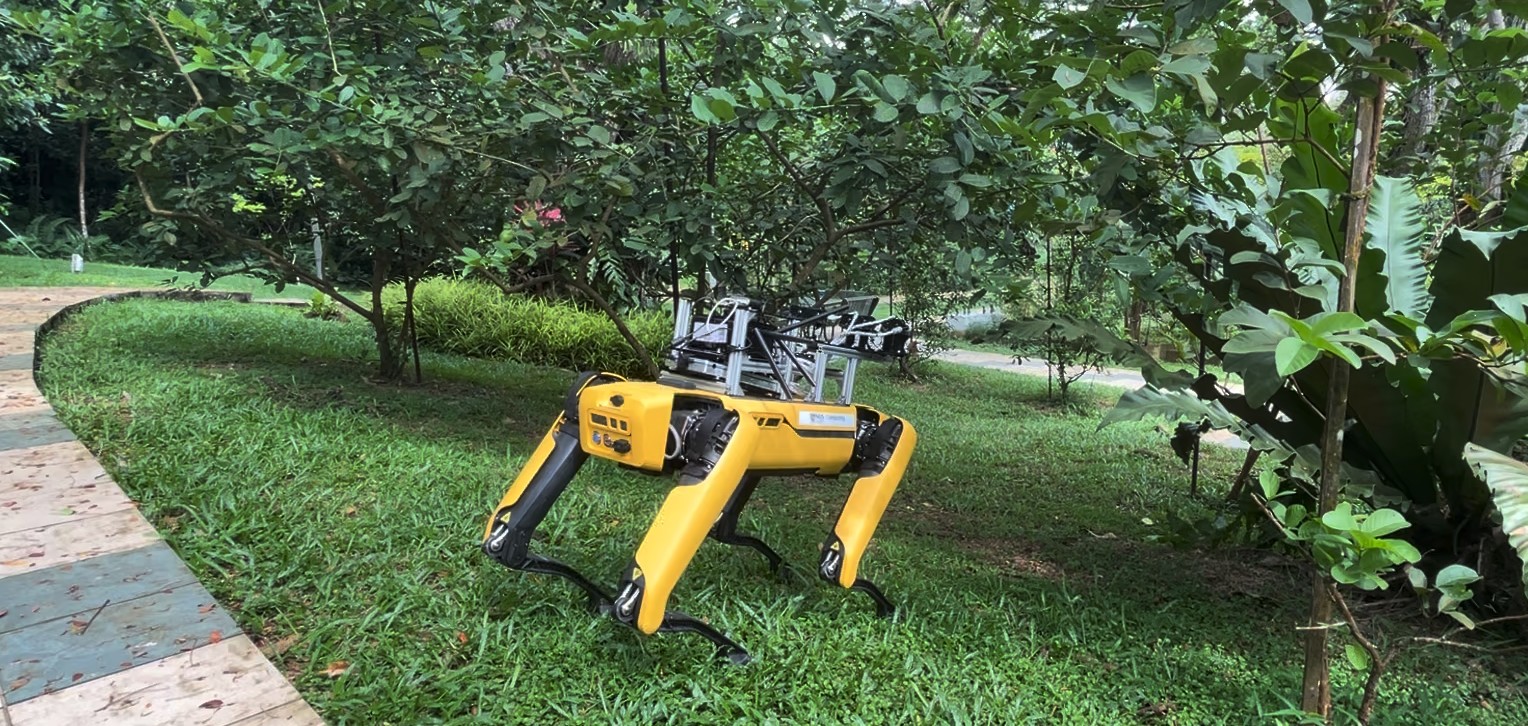} \\
         (\subfig{3}) Outdoor Pavement & (\subfig{4}) Outdoor Grassland \\
      \end{tabular} \\ 
    % (\subfig{0}) Simulation & (\subfig{1}) Lab & (\subfig{2}) Corridor & (\subfig{3}) Pavement & (\subfig{4}) Grassland \\
    \end{tabular} \vspace{-8pt}
    % \scriptsize
    \caption{Third-person view of the training and testing environments. Only data collected in \envref{0} is used to train the controller. It can be observed that the testing environment differs significantly from the simulation in terms of visual features and spatial geometry. }
    \label{fig:envs}
    \vspace{-15pt}
\end{figure*}

To quantify performance, we repeat each task $N$ times and compute success rate (\SR{}) and success weighted by path length (\SPL{}), defined as 
%%%
\begin{align*}
 \SR{} = \frac{1}{N}\sum_{i=1}^{N}s_i \quad 
 \SPL{} = \frac{1}{N}\sum_{i=1}^{N}s_i\frac{l^{*}_i}{l_i} 
\end{align*}
%%%
where $s_i$ is a boolean variable indicating success in the i-th trial, $l^{*}_i$ is the shortest path length achievable, and $l_i$ is the actual path length traversed by the controller. We evaluate all policies in five environments, with increasing levels of domain shifts (\figref{fig:envs}). Here are the baselines. \vspace{-3pt}

% To examine the above hypotheses, we use the following models as our baselines. 
\begin{enumerate}[label=\roman*., leftmargin=27pt]
    \item \rgb{}. An RGB-based policy learned from scratch.
    \item \gan{}. A policy using CycleGAN to transform real-world observations to the simulator domain, following the most recent work \cite{vr-goggles, da-bidirec}. 
    \item \maxcover{}. An agent using the max coverage feature set \cite{midlevel-navigation} as the representation.
    \item \nonav{}. An ablation model that does not convert semantics to \navigability{}. Basically, the control module $f$ directly takes in the segmentation mask and depth prediction as the input.
    \item \nosemantics{}. An ablation model with only depth as input. 
    % The control module only takes depth representation as the observation.
    \item \nodepth{}. An ablation model with only \navigability{} as input. 
    % The control module only takes \navigability{} as the observation. 
\end{enumerate}  \vspace{-5pt}
For a fair comparison, we train all controllers with the same algorithm and data.

\subsection{Implementation}

\begin{wrapfigure}{r}{0.5\textwidth}
    \vspace{-20pt}
    \includegraphics[width=0.5\textwidth]{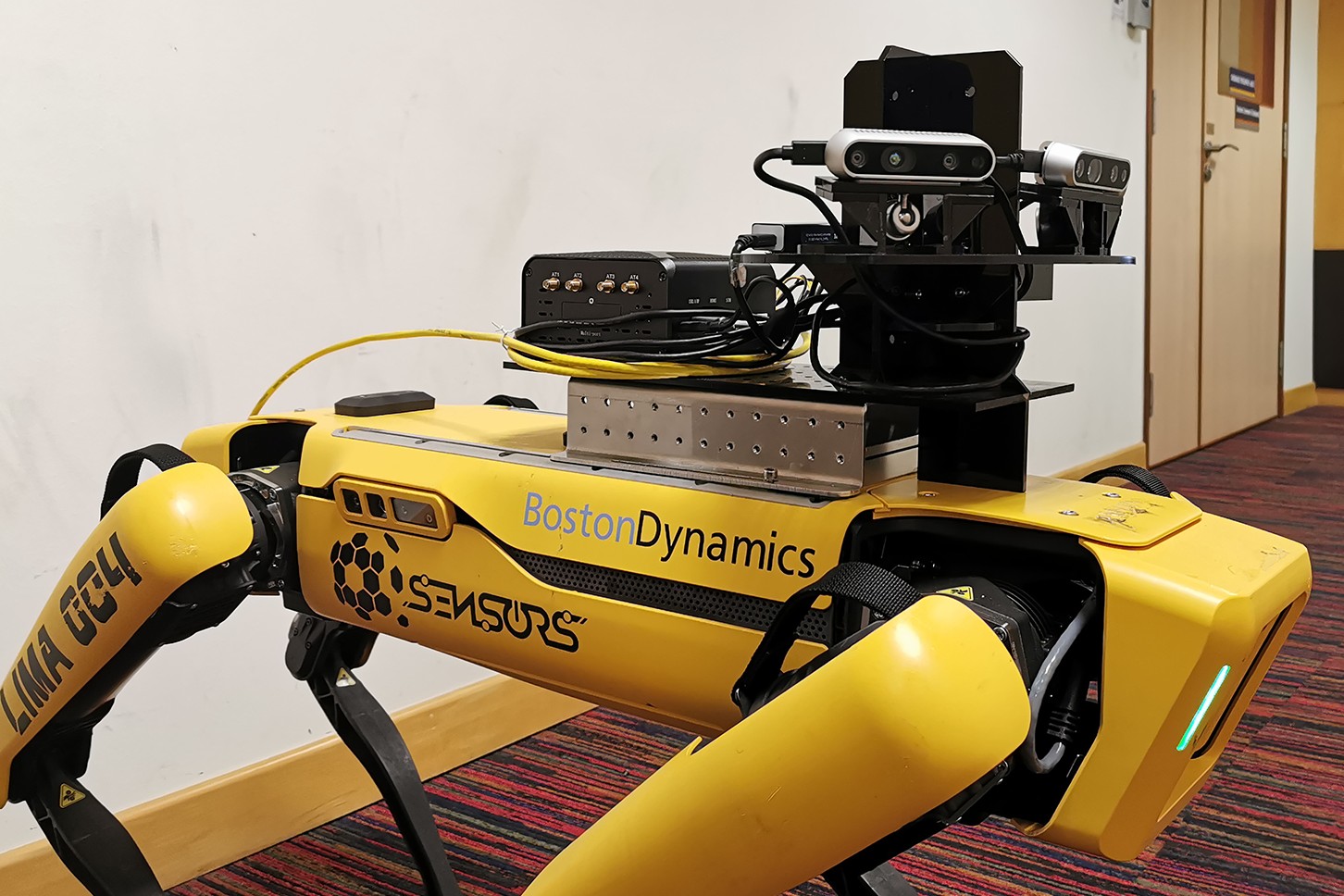}
    \caption{Real-world system setup.}
    \label{fig:hardware_system}
    \vspace{-10pt}
\end{wrapfigure} 

We use a photo-realistic indoor simulator, Habitat~\cite{savva2019habitat}, for policy learning. We use the training split in~\cite{xia2018gibson} that contains 72 maps, and we evaluate the policy in another 5 maps for simulation-based testing.  

To generate labeled data for imitation learning, we use a path planner to find the shortest path between randomly sampled initial and goal locations using the ground-truth mesh. The action space is \{\turnleft{}, \turnright{}, \goforward{}\}, as specified by the simulator. We collect 100 trajectories in each training environment. Each sample is a 4-tuple of current pose $s_t$, goal pose $s_{goal}$, RGB observation $o_t$, and the expert action $a_t$. 

We parallelly run 36 simulator processes to generate expert demonstrations, which takes 4 hours to finish and we end up having $740K$ samples. Finally, we train the policy model by minimizing the cross-entropy loss between the prediction and the expert actions. We use the AdamW optimizer \cite{adamw} with an initial learning rate of $1e\text{-}3$ and a batch size of $200$. Training is completed in approximately two days with 2 $\times$ RTX2080Ti.

% However, we do not iteratively collect data to cover the failure modes of the learned policy, \ie, performing DAGGER \cite{dagger,hg_dagger}. We would achieve a better training performance if the strategy is adopted, but this is orthogonal to this work since we only focus on reducing the performance degradation caused by domain shifts.

For real-world testing, we use the Boston Dynamics Spot robot as a physical platform, which is connected to an NVIDIA AGX Xavier onboard computer (\figref{fig:hardware_system}). We use the RGB stream of one Intel RealSense D435i RGB-D camera as the agent observation and visual odometry from the Spot robot for localization. 

% %%%
% \begin{figure}[t]
%   \centering
%   % \setlength{\tabcolsep}{1pt} % Default value: 6pt
%   % \renewcommand{\arraystretch}{0.5} % Default value: 1
%   \begin{tabular}{ccc}
%     \includegraphics[width=0.3\columnwidth]{system/spot_l.jpg} &
%     \includegraphics[width=0.3\columnwidth]{system/spot_m.jpg} & 
%     \includegraphics[width=0.3\columnwidth]{system/spot_r.jpg} \\
%   \end{tabular}
%   \caption[Hardware system setup]{Hardware system setup. }
%   \label{fig:hardware_system}
%   \vspace{-18pt}
% \end{figure}

\subsection{Results} \label{sec:midrep-gen} 

\begin{table*}[t]
\centering
\caption{Navigation experiment results. Best performance in bold. } 
  \label{tab:navigation}
% \scriptsize
 \begin{tabular}{l c c c c c c c c c c c c}
    \toprule
%----------------------------------------------------------------
    Model &  %\multicolumn{1}{c}{Model} & 
    \multicolumn{2}{c}{\thead{\envref{0} \\ Simulation \\  (N = 25)}} & 
    \multicolumn{2}{c}{\thead{\envref{1} \\  Lab \\ (N = 5)}} & 
    \multicolumn{2}{c}{\thead{\envref{2} \\ Corridor \\ (N = 5)}} &
    \multicolumn{2}{c}{\thead{\envref{3} \\ Pavement \\ (N = 5)}} &
    \multicolumn{2}{c}{\thead{\envref{4} \\ Grassland \\ (N = 5)}} &
    \multicolumn{2}{c}{\thead{Average}}  \\
    \cmidrule(lr){2-3} \cmidrule(lr){4-5} \cmidrule(lr){6-7} \cmidrule(lr){8-9} \cmidrule(lr){10-11} \cmidrule(lr){12-13}
    %     ----------------------------------------------------------------
      & \SR{} & \SPL{} & \SR{} & \SPL{} & \SR{} & \SPL{} & \SR{} & \SPL{} & \SR{} & \SPL{} & \SR{} & \SPL{} \\
      \midrule
      \rgb{} & 0.72 & 0.68 & 0 & 0 & 0 & 0 & 0.40 & 0.39 & 0.20 & 0.17 & 0.26 & 0.25 \\
      \gan{} \cite{da-bidirec} & 0.68 & 0.63 & 0.60 & 0.58 & 0.60 & 0.59 & 0.80 & 0.71 & 0.60 & 0.54 & 0.65 & 0.61 \\
      \maxcover{} \cite{midlevel-navigation} & 0.76 & 0.72 & 0 & 0 & 0.40 & 0.35 & 0.60 & 0.49 & 0.20 & 0.11 & 0.39 & 0.33 \\
      \textbf{\ours} & \textbf{0.80} & \textbf{0.76} & \textbf{1} & \textbf{0.92} & \textbf{0.80} & \textbf{0.71} & \textbf{1} & \textbf{0.94} & \textbf{0.80} & \textbf{0.64} & \textbf{0.88} & \textbf{0.79} \\
      \midrule
      \nonav{} & 0.72 & 0.67 & 0.40 & 0.37 & 0.40 & 0.32 & 0.20 & 0.17 & 0.40 & 0.37 & 0.42 & 0.38  \\
      \nodepth{} & 0.64 & 0.62 & 0.80 & 0.73 & 0.60 & 0.53 & \textbf{1} & 0.93 & 0.20 & 0.12 & 0.65 & 0.59 \\
      \nosemantics{} & 0.72 & 0.68 & 0.60 & 0.53 & 0.60 & 0.70 & 0.20 & 0.16 & 0.60 & 0.48 & 0.54 & 0.51 \\
      % \midrule 
      % Average & 0.72 & 0.68 & 0.49 & 0.45 & 0.49 & 0.46 & 0.60 & 0.54 & 0.43 & 0.35 \\
 \bottomrule
\end{tabular} \vspace{-15pt}
\end{table*}
%%%%%%%%%%%%%%%%%%%%%%%%%%%%%%%%%%%%%%%%%%%%%%%%%%%%%%%%%%%%%%%%%

% %%%
% \begin{figure}[t]
%   \centering
%   % \setlength{\tabcolsep}{1pt} % Default value: 6pt
%   % \renewcommand{\arraystretch}{0.5} % Default value: 1
%   \begin{tabular}{cc}
%     % (\subfig{a}) & \raisebox{-.5\height}{\includegraphics[width=0.8\columnwidth]{method/sim_nav_example1.png}} \\
%     (\subfig{b}) & \raisebox{-.5\height}{\includegraphics[width=0.8\columnwidth]{exp/midrep/oldlab-4feat.png}}  \\
%     (\subfig{c}) & \raisebox{-.5\height}{\includegraphics[width=0.8\columnwidth]{exp/midrep/corridor-4feat.png}} \\
%     (\subfig{d}) & \raisebox{-.5\height}{\includegraphics[width=0.8\columnwidth]{exp/midrep/pavement-4feat.png}} \\
%     (\subfig{e}) & \raisebox{-.5\height}{\includegraphics[width=0.8\columnwidth]{exp/midrep/park-4feat.png}}  \\
%   \end{tabular}
%   \caption{Representations during navigation experiments. Row (\subfig{a}) is the real-world environment, whereas wow (\subfig{b})-(\subfig{e}) corresponds to \envref{1} - \envref{4} respectively. With navigability alone, the agent suffers from size-distance ambiguity in geometrically complex environments, such as \envref{2} and \envref{4}, in which case depth information is particularly helpful for obstacle avoidance. }
%   \label{fig:exp_obs}
%   \vspace{-12pt}
% \end{figure}
% %%%

%%%

Full results are summarized in \tabref{tab:navigation}. Overall, our proposed approach outperforms all baselines across all environments. As expected, \rgb{} is unable to generalize to the real world, due to domain shifts in color, obstacle type, texture, and other variabilities. \gan{} improves upon \rgb{} by transforming the observation image back to the simulator domain, and the improvement is pronounced in real-world indoor scenes, where \rgb{} fails drastically. In outdoor scenes, it is not comparable with \ours, as it produces distorted and unnatural images when the source differs significantly from the target domain (\h{}).

% Especially in \envref{1} and \envref{2}, where there are manually placed obstacles, the \rgb{} agent collides with obstacles frequently. This is because it has difficulties identifying objects that are different from those in the simulator. 

% However, it causes a performance drop when the domain shift is mild (\envref{0}), which is because the benefits of performing image translation are outweighed by the artifacts it brings to the agent observation. In \envref{4} where the domain gap is huge, it 

Next, by comparing \ours{} with \nonav{} across different environments, we observe that the performance gap between controllers increases when the testing environment deviates from the simulation. Specifically, \ours{} maintains an \SR{} of above 0.8, but that of \nonav{} ranges from 0.2 to 0.4. This shows that converting semantic categories helps the controller generalize to semantically different scenes (\hh{}). 

% This verifies our \hh{} that the conversion is effective in bridging the semantics gap (\hh{}). 

We can also find out why we need both semantics and depth in our representation. If we compare the performance of \nodepth{}, and \nosemantics{} in \envref{3} and \envref{4}, we would find that \nodepth{} has a much better performance in \envref{4} (\SR{} 0.6 vs 0.2), whereas \nosemantics{} is significantly better in \envref{3} (\SR{} 1 vs 0.2). This is because \envref{3} is an open free space where the geometry is simple and the pavement boundary can be segmented out by semantics. On the other hand, \envref{4} has homogeneous semantics and obstacles, \ie, trees. Without geometry information, the model suffers from size-distance ambiguity. This loss of information increases the chance of collision. These show that both depth and \navigability{} are important (\hhh{}). 

Lastly, \maxcover{} uses the feature set that has the most information overlap with all visual features, among all possible combinations of the same set size (k = 4) \cite{taskonomy}. We can see it has better performance than \rgb{} in \envref{0}, \envref{2}, and \envref{3}, but not in other environments. In contrast, \ours{} has a consistent and strong performance across all environments, including outdoor scenes. This shows that our representation is informative to the task (\hhhh{}). 

Above all, we show that our representation is compact and informative, which effectively improves the generalization of the learned policy. 

% We found that this is because the representations do not improve the controller's obstacle avoidance capability in general. Specifically, all these three environments are mostly obstacle-free and have structured spatial layouts. For instance, both the navigable region boundaries in \envref{2} and \envref{3} can be easily segmented out (walls and grass respectively), but features in \envref{1} and \envref{4} are more complex. It seems that \maxcover{} is not able to deal with geometrically complex scenarios. Connecting to the fact that the only geometry-related feature in the set is the \normal{}, it is reasonable to hypothesize that (i) the \normal{} pre-trained model is not of good quality, and/or (ii) the \normal{} is not a good representation to learn obstacle avoidance skills. 

% In contrast, \ours{} has a consistent and strong performance across all environments, including outdoor scenes. Despite only having two features, \ours{} outperforms \maxcover{}, which is more sufficient in information, in both environments close to the training environment (\envref{0}) and significantly differs from the simulation. As adding more information would not improve performance, our representation is informative for the task (\hhhh{}). 

% In contrast, our model has the best of both worlds, since it incorporates both features. This verifies our \hh{} that semantics and geometry capture different aspects of the world and our model captures both. This verifies our hypothesis \hhh{}.

\subsection{Further Analysis} 

Further, we seek to understand how representations improve generalization performance. One way is to estimate the difference between domains in the representation space, and the difference can be estimated by samples, \ie, representations of the agent while operating in different environments. If some representation reduces the difference, it helps preserve the performance across environments.

However, finding such a distance measure is non-trivial. Geometry-based measures, such as L1 and Wasserstein distance, are not invariant to the numerical scale of the data. For example, simply linearly scaling RGB values to $[0, 1]$ reduces the distance between two samples, though this linear transformation should not change the domain gap. On the other hand, density-based measures, such as KL-divergence, cannot be estimated accurately with Monte Carlo methods on our high-dimensional data.

Instead, we find \adist{} \cite{generalization_theory} suitable for our purpose. We define a domain as a distribution $\D$ on an instance set $\S$. For sim-to-real transfer, we have a source domain $\mathcal{D}_S$ and a target domain $\mathcal{D}_T$. The task is to predict a label $\Z$, \ie, control, and we assume $\Z \in [0, 1]$. The ground-truth mapping from $\D$ to $\Z$ is defined as $h$. Given a representation function $\phi$ that maps $\D$ to a feature space $\Y$, the features induce distributions $\tilde{\mathcal{D}}_S$ and $\tilde{\mathcal{D}}_T$, as well as $\tilde{h}$, which is the ground-truth mapping from $\Y$ to $\Z$. Under an imitation learning setup, we learn a predictor $f$ from data, which is an approximation of $h$, then the one-step error rate of $f$ can be defined as the difference between its prediction and the output from the label-generating function $h$. Here we formalize the error rate under distribution $\mathcal{D}_S$: 
\begin{align} \label{eqn:eps_s}
    \epsilon_S(f) 
    % & = \E_{y \sim \tilde{\mathcal{D}}_S} \left[ \E_{z \sim \tilde{h}(y)} [z \neq f(y)] \right] \\
    & = \E_{y \sim \tilde{\mathcal{D}}_S} \left[ \tilde{h}(y) - f(y) \right]
\end{align}
The error under the target distribution, $\epsilon_{T}(f)$, can be defined similarly. Further, as behavior cloning suffers from the compounding error issue \cite{dagger}, the cost of rolling out policy $f$ in the target domain for $T$ steps is $\mathcal{O}(T^2\epsilon_T(f))$. 

Next, we define a learner in the hypothesis space $\H$ which performs optimally on the source and the target domains, \ie
\begin{equation}
    f^* = \argmin_{f \in \H} (\epsilon_S(h) + \epsilon_T(h))
\end{equation}
and we denote the total error rate of $f^*$ as $\lambda$, \ie, $\lambda = \epsilon_S(f^*) + \epsilon_T(f^*)$. According to \cite{generalization_theory}, we have the following generalization error bound:

\begin{theorem} \label{theorem:bound}
With probability as least $1-\delta$, for every $f$ in hypothesis space $\H$ of Vapnik-Chervonenkis (VC) dimension $d$, the following holds
    \begin{align} \label{eqn:eps_t}
        \epsilon_T(f) & \leq \hat{\epsilon}_S(f) + d_{\mathcal{H}}(\tilde{\mathcal{U}}_S, \tilde{\mathcal{U}}_T) 
        + \lambda  \nonumber \\ 
        & \quad\qquad + \frac{4}{m}\sqrt{(d\log\frac{2em}{d} + \log\frac{4}{\delta})}  
         + 4\sqrt{\frac{d\log(2m') + \log(\frac{4}{\delta})}{m'}}
    \end{align}
    where $\tilde{\mathcal{U}}_S$ and $\tilde{\mathcal{U}}_T$ are unlabelled samples of size $m'$ each drawn from $\tilde{\mathcal{D}}_S$ and $\tilde{\mathcal{D}}_T$ respectively,  $d_{\mathcal{H}}$ is the \adist{}, and $e$ is the natural logarithm base. 
\end{theorem}

The intuition behind the theory is that, the error bound under the target distribution can be decomposed into the error under the source distribution, the lowest achievable error, the domain gap, plus some terms dependent on the complexity of the function and the dataset size. For our purpose, what is important is that it explains the relation between the domain gap and model performance, and the \adist{} can be estimated from the error rate of an optimal binary classifier on the task of discriminating between samples from the two distributions. Formally, the distance is defined as
\begin{equation} \label{eqn:da}
    d_{\mathcal{A}}(\tilde{\mathcal{U}}_S, \tilde{\mathcal{U}}_T) = 2(1 - 2\min_{f' \in \mathcal{H}} err(f'))
\end{equation}
where $err$ is the binary classification error, \ie,
\begin{equation} \label{eqn:err}
    err(f') = \frac{1}{2m'} \sum^{2m'}_{i=1} |f'(z_i) - I_{z_i \in \tilde{\mathcal{U}}_S}|
\end{equation}
where $I$ is the indicator function. Note that here $\H{}$ is shared in \eqnref{eqn:eps_t} and \eqnref{eqn:da}. On our task, given that our controller $f$ is a deep neural network, it is infeasible to find $\min_{f' \in \mathcal{H}} err(f')$ in \eqnref{eqn:da}, which is a global optimum in $\H{}$. To sidestep the difficulty, we use a linear model as an approximation, \ie
\begin{align} \label{eqn:da2}
    d_{\mathcal{A}}(\tilde{\mathcal{U}}_S, \tilde{\mathcal{U}}_T) 
    & \approx 2(1 - 2\min_{f' \in \mathcal{H}'} err(f'))
\end{align}
where $\mathcal{H}'$ is the hypothesis space of the linear classifier.

\begin{table}[t]
\centering
% \scriptsize
\caption{\adist{} from the testing environment to the training environment in the representation space. }
 % \vspace{-8pt}
  \label{table:domain-gap}
 \begin{tabular}{l c c c c}
    \toprule
%----------------------------------------------------------------
    Representation &  %\multicolumn{1}{c}{Model} & 
    % \multicolumn{1}{c}{\thead{Indoor \\ Simulation}} & 
    \multicolumn{1}{c}{\thead{\envref{1} \\ Lab  }} & 
    \multicolumn{1}{c}{\thead{\envref{2} \\ Corridor }} &
    \multicolumn{1}{c}{\thead{\envref{3} \\ Pavement }} &
    \multicolumn{1}{c}{\thead{\envref{4} \\ Grassland }} \\
    % \cmidrule(lr){2-3} \cmidrule(lr){4-5} \cmidrule(lr){6-7} \cmidrule(lr){8-9} \cmidrule(lr){10-11} \\
    %     ----------------------------------------------------------------
      % & \SR{} & \SPL{} & \SR{} & \SPL{} & \SR{} & \SPL{} & \SR{} & \SPL{} & \SR{} & \SPL{} \\
      \midrule
      \rgb{} & 1.95 & 1.73 & 1.99 & 1.80 \\
      \gan{} & 1.93 & 1.62 & 1.94 & 1.74 \\
      \titlecap{\depth{}}  & \textbf{1.07} & 1.63 & 1.39 & \textbf{1.27} \\
      \titlecap{\semseg{}} & 1.24 & 1.14 & 1.38 & 1.95 \\
      \titlecap{\navigability{}} & 1.18 & \textbf{1.11} & \textbf{1.17} & 1.40 \\
      \titlecap{\ours{}} & 1.36 & 1.63 & 1.65 & 1.82 \\
      \titlecap{\unsupseg{}} & 1.61 & 1.46 & 1.88 & 1.59 \\
      \titlecap{\keypoints{}} & 1.85 & 1.61 & 1.87 & 1.67 \\
      \titlecap{\normal{}} & 1.83 & 1.77 & 1.92 & 1.90 \\
 \bottomrule
\end{tabular} \vspace{-15pt}
\end{table}

The estimated \adist{} under different representation spaces is presented in \tabref{table:domain-gap}. It can be observed that our selected representations, depth and \navigability{}, minimize the \adist{} between the target domain and the source domain among all candidates. It is interesting to see that the domain gap of \envref{1} and \envref{4} is minimized in the depth space, which is intuitive: the obstacles in these environments are cluttered and less than 5 meters away, similar to the simulated household scenes during training time. The other two environments are made close to the training environment by the navigability representation, possibly because the navigable regions are narrow with obstacles on the sides, whereas the other environments have a large open space with scattered non-navigable regions. The results are consistent with our intuitive perception of these environments. 

Comparing the domain gap in different representation spaces, we find that \rgb{} has the largest \adist{}, and \gan{} reduces the distance by a margin as expected. Comparing \semseg{} and \navigability{}, we can see that our conversion is effective in narrowing the gap, particularly in \envref{4} where the semantics are completely different from training. This is consistent with our observation from \tabref{tab:navigation} that the conversion is very effective in improving the end-task performance.

% (\ours{} 0.94 vs \nonav{} 0.17). 

The estimated \adist{} shows that our representation is compact and effectively reduces the domain gap. As pointed out by \theoref{theorem:bound}, reducing \adist{} lowers the generalization error bound, which explains the performance improvement brought by our representation. 

\section{Conclusion}

This work shows that invariant representations are key to generalization, specifically, in sim-to-real transfer for visual navigation. Our proposed approach enables a local navigation policy trained entirely in simulated indoor environments to generalize to the real world, both indoor and outdoors. We also provide quantitative analyses to show that our representation narrows the domain gap and improves the generalization performance. 

Furthermore, we expect the model to be scalable, as it leverages pre-trained models, also known as foundation models \cite{foundation-model}, to achieve zero-shot transfer to unseen domains. We believe the generalizability of the learned policy improves over time as the foundation models absorb more diverse data during pre-training. On this note, one potentially promising future direction is to explore ``foundational representations" for robotics tasks, \ie, representations that emerge from foundation models and capture the minimal sufficient information for decision-making. Having a set of such representations might enable policies to generalize seamlessly to new domains. We hope this work provides insights for future work on the broader key question of generalization in robot learning.

\paragraph{Acknowledgments} This research is supported by Agency of Science, Technology \& Research (A*STAR), Singapore  under its National Robotics Program (Award M23NBK0053).

% % %
% % % ---- Bibliography ----
% % %
\bibliography{references-camera-ready} 
\bibliographystyle{plain} 
\end{document}